\documentclass[10pt,journal,compsoc]{IEEEtran}

\usepackage{url}
\usepackage{booktabs}
\usepackage{calc}
\usepackage{graphicx}
\usepackage{soul,color}
\usepackage{multirow}
\usepackage{comment}
\usepackage{paralist}
\usepackage{color}
\usepackage{paralist}
\usepackage{xcolor}
\usepackage[nocompress,noadjust]{cite}
\usepackage{amsmath}
\usepackage{amssymb}
\usepackage{amsthm}
\usepackage{bm}
\usepackage{xspace}
\usepackage{enumitem}
\usepackage{pifont}
\usepackage[utf8]{inputenc}
\usepackage{standalone}
\usepackage{lipsum}
\usepackage[ruled, vlined, longend, linesnumbered]{algorithm2e}
\DontPrintSemicolon
\SetAlFnt{\small}
\ifCLASSOPTIONcompsoc
	\usepackage[caption=false, font=footnotesize, labelfont=sf, textfont=sf]{subfig}
\else
	\usepackage[caption=false, font=footnotesize]{subfig}
\fi

\newcommand{\R}{\mathbb{R}}
\newcommand{\N}{\mathbb{N}}

\newcommand*{\unit}[1]{\ensuremath{\mathrm{\,#1}}}

\newcommand{\descr}[1]{\medskip\noindent\textbf{#1}}
\renewcommand{\paragraph}{\descr}

\newcommand{\revision}[1]{{\color{black}{#1}}}

\DeclareMathOperator*{\argmax}{arg\,max}
\DeclareMathOperator*{\argmin}{arg\,min}

\newcommand{\mpayload}{\ensuremath{m_p}\xspace}
\newcommand{\vdrone}{\ensuremath{s_d}\xspace}
\newcommand{\fdrag}{\ensuremath{F\textsubscript{D}}\xspace}

\newcommand{\tdg}{\ensuremath{\mathcal{G}}\xspace}

\newcommand{\er}{{Erd\H{o}s--Rényi}\xspace}
\newcommand{\prob}{\textsc{MFP}\xspace} 

\newcommand{\algpp}{\textsc{PP}\xspace}
\newcommand{\algosp}{\textsc{OSP}\xspace}
\newcommand{\algdsp}{\textsc{DSP}\xspace}
\newcommand{\alggsp}{\textsc{GSP}\xspace}

\newcommand{\algpplong}{Pre-Processing\xspace}
\newcommand{\algosplong}{Offline Shortest Path\xspace}
\newcommand{\algdsplong}{Dynamic Shortest Path\xspace}
\newcommand{\alggsplong}{Greedy Shortest Path\xspace}
\newcommand{\wcu}{WCU\xspace}
\newcommand{\wcus}{WCUs\xspace}

\newcommand{\vor}{Voronoi\xspace}
\newcommand{\del}{Delaunay\xspace}
\newcommand{\hyb}{Hybrid\xspace}

\newcommand{\vgreen}{\textsc{green}\xspace}
\newcommand{\vblack}{\textsc{black}\xspace}
\newcommand{\vgray}{\textsc{gray}\xspace}

\newcommand{\scanceled}{\textsc{canceled}\xspace}
\newcommand{\ssuccess}{\textsc{success}\xspace}
\newcommand{\sdelivered}{\textsc{delivered}\xspace}
\newcommand{\sfail}{\textsc{fail}\xspace}

\begin{document}

\title{Energy-Constrained Delivery of Goods with \\ Drones Under Varying Wind Conditions}

\author{
	Francesco Betti Sorbelli,
	Federico Cor\`o,
	Sajal K.\ Das,~\IEEEmembership{Fellow,~IEEE,} and \\
	Cristina M.\ Pinotti,~\IEEEmembership{Senior~Member,~IEEE}
	

	\thanks{
	Francesco Betti Sorbelli and Sajal Das are with Dept. of Comp. Sc., Missouri Univ. of Science and Technology, Rolla, MO, USA, 65409. 
	Federico Cor\`o is with the Dept. of Comp. Sc., Sapienza Univ. of Rome, Italy.
	Cristina M.~Pinotti is with the Dept. of Comp. Sc. and Math., Univ. of Perugia, Italy.
	}
}

\IEEEoverridecommandlockouts

\IEEEtitleabstractindextext{%
\begin{abstract}
In this paper, we study the feasibility of sending drones to deliver goods from a depot to a customer by solving what we call the Mission-Feasibility Problem (MFP). Due to payload constraints, the drone can serve only one customer at a time. 
To this end, we propose a novel framework based on time-dependent cost graphs to properly model the MFP and tackle the delivery dynamics.
When the drone moves in the delivery area, the global wind may change thereby affecting the drone's energy consumption, which in turn can increase or decrease. 
This issue is addressed by designing three algorithms, namely: (i) compute the route of minimum energy once, at the beginning of the mission, (ii) dynamically reconsider the most convenient trip towards the destination, and (iii) dynamically select only the best local choice.
We evaluate the performance of our algorithms on both synthetic and real-world data. The changes in the drone's energy consumption are reflected by changes in the cost of the edges of the graphs.
The algorithms receive the new costs every time the drone flies over a new vertex, and they have no full knowledge in advance of the weights.
We compare them in terms of the percentage of missions that are completed with success (the drone delivers the goods and comes back to the depot), with delivered (the drone delivers the goods but cannot come back to the depot), and with failure (the drone neither delivers the goods nor comes back to the depot).
\end{abstract}

\begin{IEEEkeywords}
Drone delivery algorithms, energy model, wind model, time-dependent cost graphs, mission feasibility problem
\end{IEEEkeywords}
}

\IEEEpeerreviewmaketitle
\maketitle

\section{Introduction}
In recent years, drones or Unmanned Aerial Vehicles (UAVs) have been widely deployed in civil applications~\cite{shakhatreh2019unmanned}, such as agriculture, environmental protection, localization~\cite{bettisorbelli2018range, bettisorbelli2019rangefree}, and delivery of goods.  
For example, Amazon and Google are testing drone delivery systems, particularly for ``last-mile'' logistics of small items. 
There are many advantages of using drones for deliveries, including economic benefits and the ability to deliver in time-critical situations or hard-to-reach places.
Transportation companies can further extend their business relying on drones that deliver goods directly to the customer locations in medium-large size cities~\cite{rao2016societal}.

In our scenario, a transportation company is assumed to have a depot and a fleet of drones for serving the customers that are spread out in a delivery area map and can request for product deliveries.
Since the payload constraint forces the drone to go back to the depot after each delivery, the Traveling Salesman Problem is not suitable to properly model the delivery systems using UAVs; thus, it is only possible to serve each customer separately~\cite{kornatowski2018last}.
In open areas, drones can fly directly (line of sight) from the source to the destination location, thereby shortening the traveled distance and hence extending the drone's lifetime.
In high-density urban areas, drones can fly following the routes above the ground avoiding buildings.
Since by regulation drones can fly up to a maximum height depending on the landscape, we assume they fly at lower altitudes.
Flying, the energy consumption is severely influenced by the airflow~\cite{hwang2018practical}, which in turn is affected by buildings and obstacles in general. 
The drone will prefer to fly tailwind than headwind because the former requires less energy than the latter~\cite{nguyen2017extending}.
The payload weight also impacts the autonomy of the drone's battery (or energy) during its flight. 

\revision{Recently, a lot of research has been done in delivering of goods with drones~\cite{otto2018optimization}.
For such energy-constrained vehicles, the optimization of distance to travel, flight time, or energy consumption, are different goals to pursue that are aimed at increasing the number of deliveries per day.
However, dynamic and unpredictable variables (like wind) can cause unexpected battery drainage and a potential loss of drones during their flights.
Hence, it is worthy of studying algorithms that aim at optimizing the flight delivery plan of drones dealing with static and dynamic variables.}

In this paper, we investigate the delivery of goods to customers using energy-constrained drones.
Given a customer's order of goods and a drone for delivery, 
the problem is to find a feasible cycle for the drone which can be completed with its energy autonomy. 
The cycle consists of two itineraries, i.e., one depot--customer with payload and one customer--depot without payload, that can be accomplished within the available energy.
The drone's energy consumption depends on static and dynamic variables as well.
While some variables (e.g., route length, payload weight) are fixed and known before the flight, other variables (e.g., wind) may not be exactly predicted because they are exactly known only when observed, and may change several times during the flight delivery (mission).
Hence, the drone may on-the-fly decide to select different routes from those originally planned for taking advantage of the wind changes.

Our approach is to model the delivery area map with a weighted graph whose costs are time-dependent.
Since variable wind conditions affect the drone's energy consumption, 
we propose a framework to properly model such dynamicity with edge costs that change over time. 
Instead, the topology of the graph (set of vertices and edges) is static because the routes that the drone can travel on the delivery area cannot change.  
The edge costs (energies required) depend not only on the relative wind condition at the edge but also on the drone's speed and payload weight.
Under this framework, we address the problem of finding which is the percentage of flight missions that can be accomplished with success (i.e., the drone delivers the goods and returns to the depot) with a given battery energy budget.

\paragraph{Contributions.}
Our results are summarized as follows:
\begin{itemize}
	\item We define the drone Mission-Feasibility Problem (\prob), which deals with the problem of delivering goods under a given energy budget and under varying wind conditions, not fully known a priori.
	
	
	\item We design three algorithms, namely \algosplong (\algosp), \algdsplong (\algdsp), and \alggsplong (\alggsp), that statically compute the route only once, dynamically reconsider on-the-fly the most convenient trip, and dynamically select on-the-fly only the best local choice, respectively.
	
	\item We evaluate the performance of our algorithms in terms of various mission statuses, on both synthetic and quasi-real data, showing that \algosp is the most conservative approach, \algdsp completes more deliveries at a higher risk of failure, and \alggsp is the least performing.
\end{itemize}

\paragraph{Organization:}
The rest of the paper is organized as follows.
Section~\ref{sec:related} reviews related work.
Section~\ref{sec:system-model} introduces the wind and energy models, and the time-dependent cost graph representing the delivery map area.
Section~\ref{sec:problem} defines the Mission-Feasibility Problem (\prob). 
Section~\ref{sec:algorithms} proposes three algorithms for \prob whose performances are evaluated first on random graphs with synthetic data in Section~\ref{sec:simulations}, and then on a real graph with real data in Section~\ref{sec:simulations2}.
Finally, Section~\ref{sec:conclusions} offers conclusions with future directions.

\section{Related Work}\label{sec:related}
This section provides an overview of the literature on the drone delivery problem. 
We also briefly summarize the single-source shortest path algorithms for dynamic graphs. 

\subsection{Drone Delivery Problem}
\revision{The \prob 
can be considered as a particular variant of the Vehicle Routing Problem (VRP), which has been proven to be \(\mathit{NP}\)-hard~\cite{lenstra1981complexity}.
VRP asks to find a sequence of routes for a sub-set of vehicles to minimize the overall traveled distance.
Other VRP variants have been further presented, e.g., VRP with multiple trips (MT-VRP) and capacitated VRP (C-VRP), all of them \(\mathit{NP}\)-hard.
The \prob stems from both the MT-VRP and C-VRP ideas, but assumes a single vehicle (drone) independently\footnote{Even though we assume a fleet of drones, each mission is considered as \emph{separate} to others.} allowing only a single delivery at a time due to the payload constraint.
Moreover, \prob knows the routing graph, but the energy cost (i.e., weights) of the routes changes over time.}




In a recent work in~\cite{nguyen2017extending}, the problem of delivering goods with drones optimizing the energy consumption has been investigated.
In this paper, the idea is to train a black-box energy consumption model by exploring the environment and sending drones to deliver goods under varying conditions.
Initially, each segment traveled by the drone is subject to an unknown cost in terms of energy, but after many consecutive missions, the goal is to infer better routes from the previous knowledge, extending the deliveries.
Anyway, this work shows that the estimation of actual energy consumption, which depends on static and dynamic variables, is hard since modeling the interaction between drones and the environment is complex.


\revision{
A multiple-drones delivery system dealing with wind, obstacle avoidance, time windows, and maximum energy budget, is proposed in~\cite{thibbotuwawa2019planning}.
A subset of feasible deliveries is initially computed offline, and a depth-first search strategy is employed then to get the final mission plan.
The extended work in~\cite{thibbotuwawa2020uav} considers the same setting with many clusters, in which the problem is seen as a Constraint Satisfaction Problem.
In~\cite{radzki2020proactive}, solutions are seen as sequences of routing problems formulated as 0-1 knapsack problems.
However, in such works is not clear how the wind dynamism can affect the delivery planning since the wind is assumed to be constant during each mission.
Moreover, differently from our approach, each drone can perform multiple deliveries on the same mission.
}

The energy required by a drone for completing deliveries is also studied in~\cite{baek2018battery}.
Since the battery is a non-ideal power supply that transfers power depending on its state of charge, 
the prediction of the total overall time of the drone flight may be overestimated.
The authors then proposed a battery-aware delivery scheduling algorithm to accomplish more deliveries with the same battery capacity.

Special drone delivery in the context of humanitarian assistance, 
has been tackled in~\cite{rabta2018drone}. 
The objective of the model is to minimize the total traveling distance of the drone under payload and energy constraints while the recharging stations are installed to allow the extension of the drone's operating distance.
The model allows to set the constraints and then minimize the objective function with a mixed-integer linear program approach.

Finally, a drone delivery system for the last-mile logistics of small packages on mixed delivery areas modeled as Euclidean and Manhattan (or EM) grids has been introduced in~\cite{bartoli2019exact}. 
The shortest path between two destinations of an EM-grid concatenates the Euclidean and Manhattan distance metrics.
Here the drone's mission consists of performing one delivery for each destination of the grid.
The problem is to find the location for the depot that minimizes the sum of the traveled distances.
The exact solution of the problem is computed in logarithmic time in the length of the Euclidean side of the EM-grid.
A similar problem is solved in~\cite{bettisorbelli2019automated}, assuming that the drone's mission serves only a subset of the grid vertices.

\subsection{Shortest Path for Dynamic Graphs}
The first idea for solving the \prob is to compute the shortest-path from the depot to the delivery destination, and back, on a graph whose edges are labeled with the energy cost. 
There exists a rich body of literature on algorithms developed for single-source shortest path problem, especially for topology dynamic graphs.
In~\cite{ramalingam1996computational}, an adaptation of Dijkstra's algorithm in the case of edge deletion or insertion is proposed.
The extended work in~\cite{frigioni2000fully} considers the change of edge costs.
In~\cite{narvaez2000new,narvaez2001new} different algorithms are proposed adapting the classic static algorithms for the single-source shortest path due to Dijkstra, Bellman-Ford, and D'Esopo-Pape, and follow an interpretation of the Ball-and-String model.
In~\cite{eramo2008design}, a new dynamic routing algorithm is proposed using multi-path information (i.e., evaluating all possible shortest paths) to quickly compute the new shortest paths in case of edge additions or deletions. 
We refer to~\cite{nannicini2008shortest} for a survey on the single-source shortest path problem on graphs whose topology and cost change over time. 

To the best of our knowledge, the online approach for the shortest path problem has been pursued mainly for the stochastic shortest path problem. 
Such a problem has been long studied in the machine learning community, for example, in the framework of {\em adversarial bandit problem}~\cite{cesa2012combinatorial}
using a Markov decision problem where an agent moves in an acyclic graph with random transitions.
This context is completely different from ours: the reward seems local, the graph is acyclic, and the weights are stochastic.

In the next section, we introduce some pre-requisites for the definition of the \prob.

\section{System Model}\label{sec:system-model}
This section introduces the relative wind and energy models for measuring the energy spent by the drone while traversing the delivery route.
We also define the time-dependent cost graph that represents the possible drone movements, along with their energy consumption, on the delivery map.

\subsection{\revision{The Relative Wind Model}}\label{sec:wind}
\begin{figure}[htbp]
    \centering
    \hfill
	\subfloat[$\omega_o(e)$.]{%
		\includegraphics[scale=0.9]{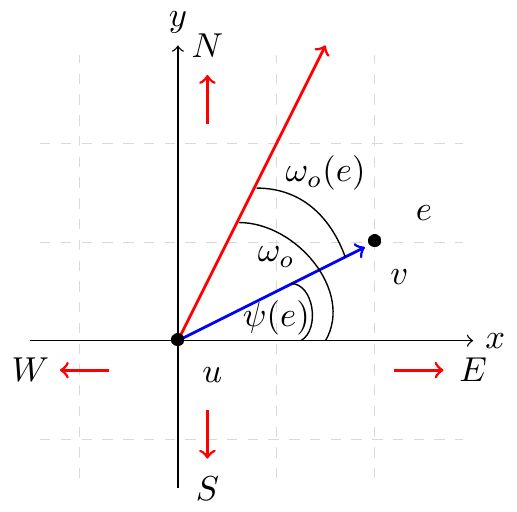}
		\label{fig:wind-direction}
	}
	\subfloat[$\omega_o(e')$.]{%
		\includegraphics[scale=0.9]{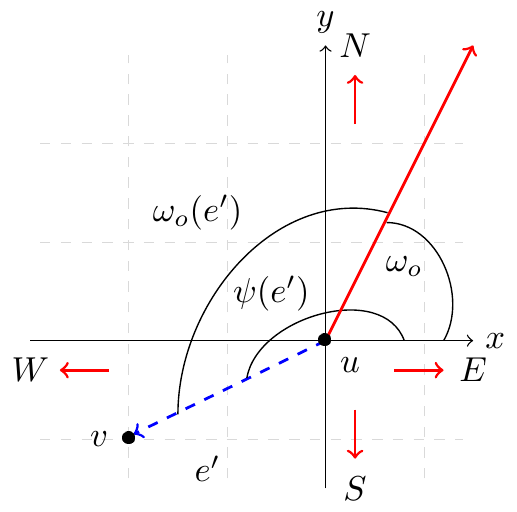}
		\label{fig:wind-direction2}
	}
	\vspace{-0.3in}
	\hfill
	\subfloat[Relative wind direction classes.]{%
		\includegraphics[scale=0.9]{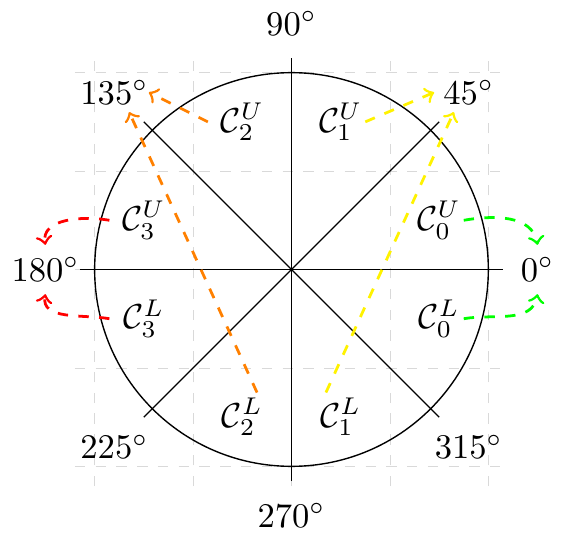}
		\label{fig:wind-classes}
	}
	\caption{Relative wind directions of an edge $e$ (solid blue (a)) and $e'$ (dashed blue (b)), and the global wind direction (red). Note $\psi(e')=\psi(e) + 180^{\circ}$. For each class, the representative angle is highlighted (c).}
	\label{fig:winds}
\end{figure}

Let the {\em global wind} $\omega = (\omega_s, \omega_o)$ be the wind in the delivery area, where $\omega_s$ and $\omega_o$ are, respectively, the speed and the direction (orientation) of $\omega$.
To define the {\em relative wind direction} $\omega_o(e)$ experienced by the drone while traversing edge $e=(u,v)$, we build a Cartesian coordinate system with origin in $u$ (an example is depicted in Figure~\ref{fig:wind-direction}).
Let $-90^{\circ} \le  \arctan(x) < 90^{\circ}$, for any $x \in \R$.
On the straight line $e$ traversed by the drone from $u=(x_u,y_u)$ towards $v=(x_v,y_v)$, the relative wind direction is $\omega_o(e)=\omega_o -\psi(e)$, where: 
\begin{equation*}
    \psi(e) = 
    \begin{cases}
        \arctan(\frac{y_v}{x_v}) \mod 360^{\circ} & \text{if $x_v > 0$} \\
        180^{\circ} + \arctan(\frac{y_v}{x_v}) & \text{if $x_v < 0$}
    \end{cases}
\end{equation*}
is the angle direction of edge $e$. 
Flying along routes with different angle directions, the drone experiences different relative wind directions.
Note that, $\psi(e)$ and $\psi(e')$ (see Figure~\ref{fig:wind-direction2}) differ by $180^{\circ}$ because $\omega_o(e')=\omega_o - \psi(e')= \omega_o -\psi(e) - 180^{\circ}= \omega_o(e) -180^{\circ}$. 

Following the weather reports as in the newspapers, in this paper we simplify the occurrences of the relative wind direction
by grouping the values of $\omega_o(e)$ in $8$ sectors (see Figure~\ref{fig:wind-classes}).
For reasons that will become clear in the next section,
we group the sectors two-by-two in $4$  classes $\mathcal{C}_0, \mathcal{C}_1, \mathcal{C}_2$, and $\mathcal{C}_3$ 
as follows:
$\mathcal{C}_i = \mathcal{C}_i^U \cup \mathcal{C}_i^L$, where
$\mathcal{C}_i^U = [iL, (i+1)L[$ and $\mathcal{C}_i^L = ](7-i)L, (8-i)L]$, for $i=0, \ldots 3$, where $L=\frac{360^{\circ}}{8}$.
For example, the class $\mathcal{C}_0$ contains the angles in $\mathcal{C}_0^U = [0^{\circ}, 45^{\circ}[$ 
and in $\mathcal{C}_0^L = ]315^{\circ}, 0^{\circ}]$.
For each class $i$, we select as the representative relative wind the angle 
with maximum cosine in absolute value, i.e., $\argmax_{\theta \in \mathcal{C}_i}{|\cos(\theta)|}$.
So, $\mathcal{C}_0$ will be represented by $0^{\circ}$,
$\mathcal{C}_1$ by $45^{\circ}$,
$\mathcal{C}_2$ by $135^{\circ}$, and
$\mathcal{C}_3$ by $180^{\circ}$.


\subsection{The Energy Model}\label{sec:energy}
The motion of drones is regulated by physical properties, such as the gravity and forces due to forward motion and wind~\cite{hua2009control}. 
The total required thrust is $T = W g + \fdrag$,
where $W$ is the total weight of the drone which includes the {\em payload weight} (mass) \mpayload, $g=9.81\unit{m/s^2}$ is the gravitational constant, and $\fdrag = \frac{1}{2} \rho \, {s_a(e)}^2 \, C_D A$ is the total drag force,
where $\rho$ is the air density, $s_a(e)$ is the drone's relative air speed, $A=\pi R^2$ is the cross sectional area ($R$ is the rotor radius), and $C_D$ is the drag coefficient.
The air speed $s_a(e)$ depends on the global wind speed $\omega_s$ and the relative wind direction $\omega_o(e)$.
The air speed $s_a(e)$  can be calculated as:
\begin{equation}
    \label{eq:air-speed}
    s_a(e) = \sqrt{s_N^2 + s_E^2},
\end{equation}
where $s_N = \vdrone - \omega_s \cos(\omega_o(e))$ and $s_E = - \omega_s \sin(\omega_o(e))$,
while $\vdrone$ is the {\em drone speed}.
Note that when $\omega_s=0$, i.e., no wind, $s_a(e)$ is the same for every drone direction $e$.
Instead, if $\omega_s\not =0$, when $\omega_o(e)=0^{\circ}$, i.e., when the global wind direction and the drone direction have the same orientation, the East component $s_E$ of the friction is null, and the North component $s_N$ has the minimum value, implying $s_a(e)$ is minimum. 
Also, when $\omega_o(e)=180^{\circ}$, i.e., when wind and drone direction are opposite, $s_E$ is null, and $s_N$ is maximum, and thus $s_a(e)$ is maximum.
Note that the $2$ sectors grouped in a single class in Section~\ref{sec:wind} span the same air speed values. Namely, since the angles of the $2$ grouped sectors share the same cosine and have opposite sine, they return the same value in Eq.~\eqref{eq:air-speed}.

Having computed the value of the total thrust $T$, it is possible to estimate the required power $P$ for a steady flight, which is $P = T(\vdrone \sin(\alpha) + s_i)$,
where $\alpha = \arctan{\left( \frac{\fdrag}{W g} \right)} $
is the pitch angle~\cite{johnson2012helicopter}, and $s_i$ is the induced velocity required for a given thrust $T$. 
Then, $s_i$ can be obtained by solving the implicit equation~\cite{johnson2012helicopter}:
\begin{equation*}
	s_i = \frac{s_h^2}{\sqrt{(\vdrone \cos(\alpha))^2 + (\vdrone \sin(\alpha) + s_i)^2}},
\end{equation*}
where $s_h = \sqrt{\frac{T}{2 \rho A}}$ is the induced velocity at hover~\cite{johnson2012helicopter}:
Note that, even not made explicit in the notation, from our description, $\fdrag$ and so $\alpha$ and $P$ depend on the drone direction $e$.
Hence, fixed a global wind $\omega=(\omega_s,\omega_o)$, the energy efficiency (or, unitary energy) $\mu(e)$ of travel along a segment $e$ is calculated as the ratio between power consumption $P$ and average ground speed $\vdrone$ of the drone, i.e., 
\begin{equation}\label{eq:unitary-energy}
    \mu_{\omega}(e) = {P}/{\vdrone}.
\end{equation}
Therefore, fixed a global wind $\omega=(\omega_s,\omega_o)$, the energy consumed for traversing one edge $e$ of length $\lambda(e)$ can be expressed as:
\begin{equation}
    \label{eq:energy}
    d_{\omega}(e) = \mu_{\omega}(e) \lambda(e).
\end{equation}

\begin{figure}[htbp]
	\subfloat[$\mpayload=7 \unit{kg}$.]{%
		\includegraphics[scale=0.75]{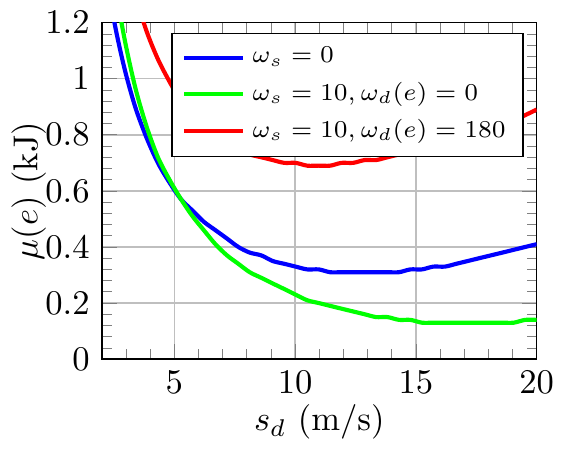}
		\label{fig:speeds_m7}
	}
	\subfloat[$\mpayload=0 \unit{kg}$.]{%
		\includegraphics[scale=0.75]{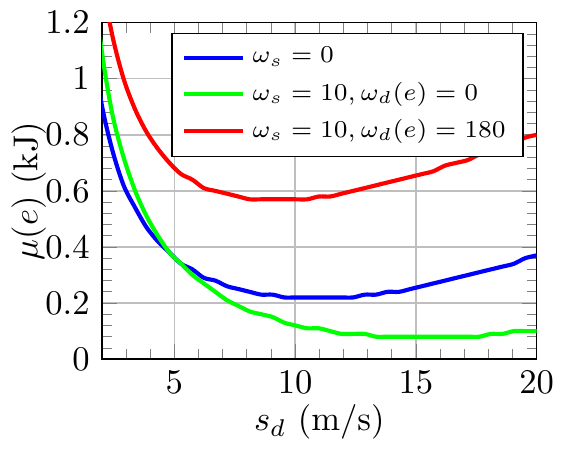}
		\label{fig:speeds_m0}
	}
	\caption{\revision{The unitary energy consumption $\mu_{\omega}(e)$ when the average speed $\vdrone$ varies, having
	fixed the payload.}}
	\label{fig:comparison_physical_model}
\end{figure}

For completeness, Figure~\ref{fig:comparison_physical_model} plots the energy spent to traverse an edge of unitary length by an octocopter (a suitable drone for delivering using proper parameters~\cite{stolaroff2018energy}) under varying wind conditions, payloads, and the drone speed.
More weight implies more energy consumption, and tail (head) wind represents less (more) energy drainage.

\subsection{\revision{Time-Dependent Delivery Network Graph}}\label{sec:graph}
To better characterize the dynamicity of the wind over the delivery area map in which the drone has to accomplish deliveries, we rely on \emph{time-dependent cost graphs}~\cite{wang2019time}.
In such graphs, the weights associated with edges dynamically change over time.
Formally, a time-dependent graph is defined as $G_t=(V, E; d_t)$, where $V$ is the set of vertices, $E$ is the set of edges, and $d_t: E \rightarrow \R^+$ is the edge cost function which represents the energy spent for traversing the edge $e \in E$ at time \(t\), where \(t\) is a time variable in a discrete time domain, i.e., \(t \in \N\).
Note that the discrete-time assumption is reasonable since in a real-world scenario the weather is usually checked at regular intervals of time~\cite{wu2007literature}. 
We refer to such regular intervals as {\em time-slot}.

The vertex set $V=\{ v_0, v_1, \ldots, v_n\}$ represents points in the plane where there are \emph{wind control units} (\wcus).
We assume that the \wcus work synchronously and thus share the same time $t$.
For simplicity, we also assume that the customers (i.e., the delivery points) are located at the graph vertices.
Vertex $v_0$ is the base of the depot from where the drone starts (and finishes) any delivery task. 
Any directed edge $e=(v_i,v_j) \in E$, for \(i \neq j\), represents the route traversed by the drone to go from vertex $v_i$ to $v_j$. 
We remark that the set of vertices $V$ and edges $E$ remain unchanged over time, only the observed global wind may change.
At every discrete time $t$, the global wind $\omega^t$ is observed. 
Thus, according to Eq.~\eqref{eq:energy}, to traverse a directed edge \(e \in E\) starting at time $t$,
the drone consumes:
$d_{\omega^t}(e)=\mu_{\omega^t}(e)\lambda(e)$. With little abuse of notation, from now on, we will simply write
\(d_t(e)=\mu_t(e)\lambda(e)\).
In the remainder of this paper, we will refer to the $t^{th}$ snapshot \(G_t\) as the graph $G=(V,E)$ whose edges are labeled $d_t(e)$, for $t \in \N$.

For any pair of vertices \(v_i, v_j \in V\), let $\pi_{ij}$ denote a simple path (i.e., a sequence of vertices without repetitions) from \(v_i\) to \(v_j\).  
At time $t$,  the energy cost of traversing the path $\pi_{ij}$ can be pre-computed assuming the cost given in $G_t$.
That is, $d_t(\pi_{ij})= \sum_{e \in \pi_{ij}} d_t(e)$. 
However, this is not the actual cost of traversing $\pi_{ij}$ starting at time $t$.
In fact, every time the drone reaches a vertex, the cost of the edges is refreshed according to the current snapshot, and the cost of $\pi_{ij}$ can be different. 
Specifically, assuming $\pi_{i_0i_z}$ formed by $z$ edges, i.e., $\pi_{ij}= (v_{i_0},v_{i_1}), (v_{i_1},v_{i_2}), \ldots, (v_{i_{z-1}},v_{i_z})$ and letting $t_0$ be the time the drone starts to traverse the first edge $(v_{i_0},v_{i_1})$, the {\em actual cost} $\tilde{d}_{t_0}(\pi_{ij}) = d_{t_0}(v_{i_0},v_{i_1}) + d_{t_1}(v_{i_1},v_{i_2}) + \ldots + d_{t_{z-1}}(v_{i_{z-1}},v_{i_z})$, where the subscript $t_j$, with $0 \le j \le {z-1}$, denotes the discrete time when the drone flies from the (intermediate) vertex $v_j$, i.e., the snapshot observed in $v_j$, from which the  cost of the edge $(v_{i_{j-1}},v_{i_j})$ derives. 
Say it differently, $(t_j-t_{j-1})$ is the actual number of time-slots required by the drone to traverse the edge $(v_{i_{j-1}},v_{i_j})$.
Clearly, $t_0 < \ldots < t_{z-1}$.
We remark that the drone cannot insert any delay during the flight.

Let $\pi^*_{ij}$ be a path with the minimum actual cost $\tilde{d}_{t_0}(\pi^*_{ij})$ between $v_i$ and $v_j$ starting at time $t_0$.
Clearly, $\pi^*_{ij}$ depends on a sequence of snapshots and can be fully computed at time $t_0$ only assuming full knowledge of all the future snapshots.
Finally, we will denote this sequence of observed graphs at different time-slots as the {\em time-dependent cost delivery network graph}, \tdg.

\section{\revision{Mission-Feasibility Problem}}\label{sec:problem}
Given a delivery area represented by a graph $G=(V,E)$ and the observed global wind $\omega^t$ at time $t$, given a budget \(B\) that represents the full charge of the drone's battery, the payload \mpayload, the drone's speed \vdrone, and selected a vertex $v_c$ as a customer's destination, the {\em Mission-Feasibility Problem} (\prob) of delivering goods at $v_c$ with a single drone aims to establish if the {\em mission}, i.e., a complete back and forth from the drone's depot to $v_c$, is feasible. 
That is, \prob aims to find if there is a suitable cycle \(C\) in $G=(V,E)$ formed by $2$ consecutive itineraries, one which starts at the source vertex $v_0$ (the depot) at time $t \ge 0$ (denoted by $\pi_{0c}$), and one which starts at the destination vertex $v_c$ (the customer) at time $t_f > t$ (denoted by $\pi_{c0}$) going back to $v_0$, whose actual energy cost $\tilde{d}_{t}(C)=\tilde{d}_{t}(\pi_{0c})+\tilde{d}_{t_f}(\pi_{c0}) \le B$.
Note that $t_f$ is not known at the time $t$.
Moreover, the departing and returning itineraries have different payloads because, after visiting $v_c$, the packages have been delivered, and the drone returns empty (i.e., $\mpayload=0$). 
We also assume that when the drone accepts a mission, it is forbidden to land at intermediate vertices and wait there for some time.
We also assume that we do not have knowledge of the full $\tdg=\{G_0, G_1,\ldots\}$, but they will be discovered online.


At the starting time $t$, the snapshot $G_t$ is available.
If the edge costs do not change over the mission, i.e., the snapshots remain the same, thus the energy cost of the edges remains unchanged during the whole mission. 
Then, the \prob has an optimal polynomial-time solution computed at the starting of the mission. 
Indeed, this variant of \prob is strictly related to the Shortest Path Problem by labeling each edge $e$ with the initial energy cost $d_t(e)$.
The minimum cost cycle $C^*$ is obtained by concatenating the two minimum energy cost paths $\pi^*_{0c}$ and $\pi^*_{c0}$ under different payloads (the first with payload, the second without), thus obtaining a cycle of cost $d(C^*)=d(\pi^*_{0c})+d(\pi^*_{c0})$, where we omit $t$ because the costs do not change over time.  
Then, the mission towards destination $v_c$ is feasible if and only if $d(C^*) \le B$. 

\begin{figure}[htbp]
	\centering
	\null\hfill
	\subfloat[$G_t$ for $t=0,1,2$.]{%
		\includegraphics[scale=0.85]{{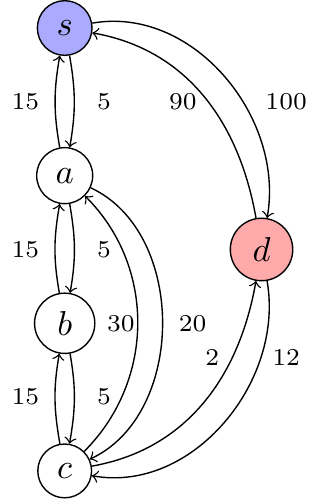}}
		\label{fig:graph-example3.1}
	}
	\hfill
	\subfloat[$G_t$ for $t \ge 3$.]{%
		\includegraphics[scale=0.85]{{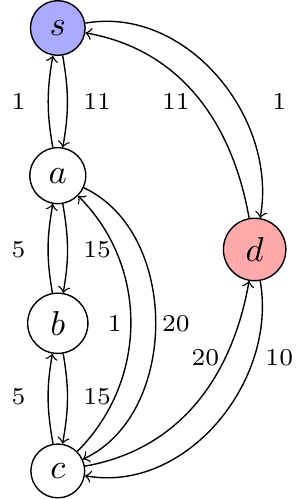}}
		\label{fig:graph-example3.2}
	}
	\hfill\null
	\caption{A \tdg whose optimal solution is not a simple cycle. 
	Edge costs for snapshots $G_0, G_1, G_2$ are in (a), while for $G_3, G_4, \ldots,$ are in (b). 
	Depot $s$ in blue, destination $d$ in red.}
	\label{fig:graph-example3}
\end{figure}

In general, however, the edge costs can change over time, and the itinerary computed as the shortest path using the edge cost of snapshot $G_t$, may become infeasible during the mission. 
The problem becomes much more complex, and it is no longer the shortest-path on a single weighted graph. Even assuming to know in advance all the snapshots in $\tdg$ (and thus the full knowledge of the winds and edge costs), the problem of finding the minimum energy-cost is more complex than computing the shortest path.
For example, the minimum energy-cost path from $v_0$ to $v_c$ with full knowledge of $\tdg$ is not always a simple path, as illustrated in Figure~\ref{fig:graph-example3}. 
In this example, we assume that the time to traverse the long edges $(c,a), (a,c), (s,d), (d,s), (c,d),$ and $(d,c)$ is equal to $2$ time-slots, while $1$ time-slot is required for the other edges.
Hence, the energy cost for traversing the edge, say,  $(v_i,v_j)$ is defined by the edge cost of the reference snapshot. 
Under these conditions, the minimum cost for the mission that starts at time $0$ (i.e., $G_0$) and has depot $s$ and destination $d$ is given by the path $(s,a,b,c,a,s,d)$ which is not a simple path.
Namely, starting at time $0$, the shortest path from $s$ to $d$ is $(s,a,b,c,d)$, since the costs remain unchanged up to time $2$ (see Figure~\ref{fig:graph-example3.1}) when the drone reaches vertex $c$. At time $3$, the costs follow snapshot $G_3$ and change as shown in Figure~\ref{fig:graph-example3.2}.
Observe that the shortest path from $c$ to $d$ is $(c,a,s,d)$ whose cost, including the cost of $(s,a,b,c)$, is $18$.
The return to $s$ has cost $11$, and thus the final cost is $18+11=29$. \prob is feasible if $B \ge 29$.

\begin{figure}[htbp]
	\centering
	\null\hfill
	\subfloat[$G_{2i}$, for $i\ge 0$.]{%
		\includegraphics[scale=0.85]{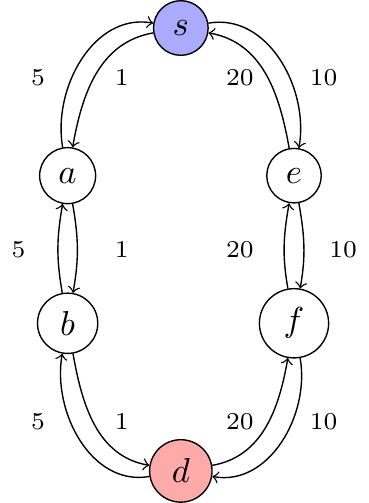}
		\label{fig:graph-example4.1}
	}
	\hfill
	\subfloat[$G_{2i+1}$, for $i\ge 0$.]{%
		\includegraphics[scale=0.85]{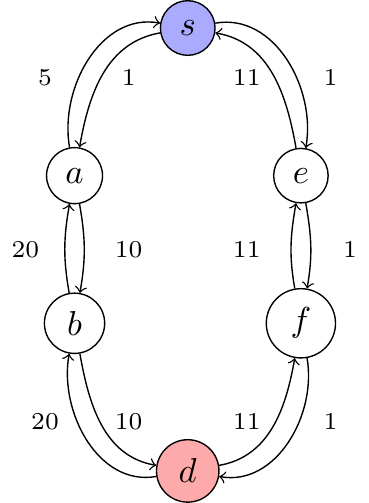}
		\label{fig:graph-example4.2}
	}
	\hfill\null
	\caption{A \tdg that does not admit a feasible solution from $s$ to $d$. 
	Edges costs for the ``even'' graphs $G_0, G_2, G_4$ are in (a), and for the ``odd'' graphs $G_1, G_3, G_5$ are in (b). 
	Depot $s$ in blue, destination $d$ in red.}
	\label{fig:graph-example4}
\end{figure}

Let us now consider another example. 
Consider $\tdg$ obtained by alternating between the graphs in Figure~\ref{fig:graph-example4}.
In this example ($s$ depot, $d$ destination), we consider all the edges of the same length, and $1$ time-slot is required to traverse every edge. 
In other words, the drone observes the costs in Figure~\ref{fig:graph-example4.1} and Figure~\ref{fig:graph-example4.2}, respectively, at snapshot $G_{2i}$ and $G_{2i+1}$, where $i \ge 0$.
In this case, with full knowledge of \tdg, the best solution of \prob has cost $42$ by traversing $(s,a,b,d,b,a,s)$.
However, with no full knowledge of the snapshots, if the drone decides to recompute the shortest-path at every new snapshot, at time $0$ the drone moves to $a$ because the shortest path in $G_0$ is $(s,a,b,d)$. 
At time $1$ when the drone has reached $a$, the shortest path according to the snapshot $G_1$ from $a$ to $d$ is $(a,s,e,f,d)$, so the drone moves back to $s$. 
By toggling back and forth between $G_0$ and $G_1$, the drone will dissipate its energy budget (no matter how large $B$ is) without ever reaching the destination.

\subsection{\prob Pre-Processing}
In this section, we analyze the feasibility of the vertices introducing the \algpplong (\algpp) Algorithm.
Given the delivery area represented by the graph $G= (V, E)$ and the drone budget $B$, the goal of \algpp is to classify the vertices according to their feasibility with respect to the wind conditions:
\begin{inparaenum}[(i)]
	\item vertices in \vblack color bring a mission to a failure, i.e., the drone will never be able to deliver the package nor return to the depot whatever are the wind conditions, and hence the mission is canceled in the beginning;
	\item vertices in \vgreen color always allow the drone to safely accomplish its mission whatever are the wind conditions; and
    \item vertices in \vgray color imply the mission completion is not guaranteed a priori and depends on the wind conditions.
\end{inparaenum}

Let $\epsilon_m=\min \{\mu(e)\}$ and $\epsilon_M=\max \{\mu(e)\}$ be, respectively, the lower and upper bound of the  unitary energy cost (see Eq.~\eqref{eq:unitary-energy}) among all possible values of $\omega$.
Note that, both the bounds can be computed because we assume a maximum wind speed and other bounded parameters (we detailed such values in Section~\ref{sec:simulations}).
Hence, we define the \emph{lower bound graph} $G^L=(V, E^L)$ where $\forall e \in E^L, d(e) = \epsilon_m \lambda(e)$, and the \emph{upper bound graph} $G^U=(V, E^U)$ where $\forall e \in E^U, d(e) = \epsilon_M \lambda(e)$.
Then, computing the shortest cycle, i.e., the shortest paths $\pi^*_{0c}$ and $\pi^*_{c0}, \forall v_c \in V \setminus \{v_0\}$ on both the static graphs $G^L$ and $G^U$, we determine the vertices \vblack for which the mission is never feasible, and also the vertices \vgreen for which \prob solution is always feasible, whatever the mission parameters are.

\begin{algorithm}[ht]
    $G^L \gets \textsc{lowerBoundGraph}(G, \epsilon_m)$\;\label{code:algpp-eworst}
    $G^U \gets \textsc{upperBoundGraph}(G, \epsilon_M)$\;\label{code:algpp-ebest}
	\ForEach{$v_c\in V \setminus \{v_0\}$}{\label{code:algpp-foreach}
		$v_c.col \gets \vgray$\;\label{code:algpp-gray}
		
		$C \gets \textsc{shortestCycle}(G^U, v_0, v_c)$\;\label{code:algpp-shortestcycle}
		\lIf{$ d(C) \le B$} {\label{code:algpp-check1}
			$v_c.col \gets \vgreen$\label{code:algpp-white}
		}
		
		$C \gets \textsc{shortestCycle}(G^L, v_0, v_c)$\;
		\lIf{$ d(C) > B$} {\label{code:algpp-check2}
			$v_c.col \gets \vblack$\label{code:algpp-black}
		}
	}
	
	\caption{\algpp\((G, B)\)}
	\label{alg:algpp}
\end{algorithm}

Algorithm \algpp, reported in Algorithm~\ref{alg:algpp}, works as follows. 
First build $G^L$ and $G^U$, and then determine the minimum cost cycle $C$ from/to the source $v_0$ after visiting the destination $v_c$ and taking into account the vertex set $V$ and the new edge set $E^U$ (Line~\ref{code:algpp-shortestcycle}).
The goal is to check if such a cycle $C$, considered in the worst-case scenario, has a cost $d(C) \le B$.
If so, the drone can safely accomplish the mission and the vertex $v_c$ is \vgreen (Line~\ref{code:algpp-white}).
The same policy can be repeated in a symmetric way considering the best case scenario and assuming the strongest tailwind computing the edge set $E^L$ of the new graph (Line~\ref{code:algpp-ebest}).
In this case, if the estimated energy consumption is larger than the budget, i.e., $d(C) > B$, the vertex is \vblack (Line~\ref{code:algpp-black}).
All the vertices that do not fall in the above two categories are \vgray.

\section{Drone Delivery Algorithms}\label{sec:algorithms}
In this section, we solve the \prob when the destination is a  \vgray vertex (the solution is trivial when a vertex is  \vgreen or  \vblack). 
In particular, we describe three new algorithms, called \algosplong (\algosp),  \algdsplong (\algdsp), and \alggsplong (\alggsp).

\begin{figure}[htbp]
	\hfill
	\subfloat[]{%
		\includegraphics[scale=0.85]{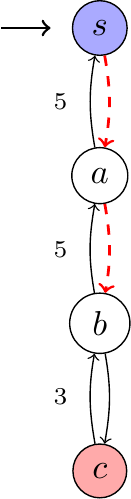}
		\label{fig:sabort}
	}
	\hfill
	\subfloat[]{%
		\includegraphics[scale=0.85]{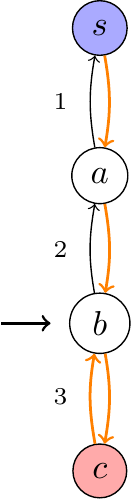}
		\label{fig:sdelivered}
	}
	\hfill
	\subfloat[]{%
		\includegraphics[scale=0.85]{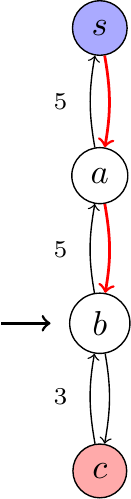}
		\label{fig:sfail}
	}
	\hfill
	\subfloat[]{%
		\includegraphics[scale=0.85]{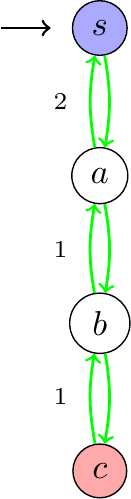}
		\label{fig:ssuccess}
	}
	\hfill\null
	\caption{\revision{Statuses when $B = 10$. Arrow is the drone's position. Dashed and solid colored lines represent planned (not traveled) and traveled edges, respectively.
	Depot $s$ in blue, destination $d$ in red.
	In detail: \scanceled (only \algosp) in (a); \sdelivered in (b); \sfail in (c); \ssuccess in (d).}}
	\label{fig:statuses}
\end{figure}

Noting that the usual solution for the shortest path is a simple path, hence we aim at finding a feasible cycle consisting of two simple paths, one going towards the destination $v_c$ and another returning to the source $v_0$. 
In other words, our solutions forbid the route to visit the same vertex twice on both the onward and return path, while these two paths may intersect at a common vertex. 
The rationale behind this requirement is to avoid the situations illustrated in Figures~\ref{fig:graph-example3}--\ref{fig:graph-example4}, and limiting the total number of edges in these cycles to $2(|V|-1)$.
Applying the \prob to a \vgray vertex, the proposed algorithms can output the following statuses:
\revision{
\begin{itemize}
    \item {\bf \scanceled (\textsc{C}):}
    corresponds to a mission in which the drone abandons to fly from the beginning because the expected energy consumption is larger than the capacity of the battery.
    Such estimation is computed considering the current weight of the graph $G_0$.
    Only the \algosp algorithm can return such status.
    

    \item {\bf \sdelivered (\textsc{D}):} 
    corresponds to a started mission in which the drone only reaches the delivery destination, but not the depot.
    This means that the drone was not able to completely accomplish the mission.
    We consider this status as a ``half success''.
    All the proposed online algorithms return such status.

    \item {\bf \sfail (\textsc{F}):} 
    corresponds to a started mission in which the drone even fails to reach the first delivery destination. 
    This happens due to the unexpected battery drainage.
    All the online algorithms return such status.
    
    \item {\bf \ssuccess (\textsc{S}):} 
    corresponds to a started mission in which the drone serves the customer and successfully gets back to the depot. That is, the actual energy cost is no more than $B$.
    All the algorithms return such status.
\end{itemize}
In Figure~\ref{fig:statuses} we report numerical examples of the aforementioned statuses.
For the sake of simplicity, we assume a straight path and same costs for the edges on each snapshot.
}

\subsection{The \algosp Algorithm}
The \algosplong (\algosp) algorithm, reported in Algorithm~\ref{alg:algosp}, computes a cycle $C$ considering the cost function as input at \revision{\(G_0\), when the mission starts at the source}.

\begin{algorithm}[ht]
    $S \gets \emptyset, flag \gets \textbf{false}, t \gets 0$\;\label{code:algosp-status} 
	$C \gets \textsc{shortestCycle}(G_0, v_0, v_c)$\;\label{code:algosp-shortestcycle}
	\If{$ d_0(C) \le B$} {\label{code:algosp-check1} 
	    \ForEach{$e=(v_i,v_j) \in C$}{\label{code:algosp-loop} 
	        $B \gets B - d_t(e)$\;\label{code:algosp-budgetupdate} 
	        \lIf{$v_j = v_c$}{
	            $flag \gets \textbf{true}$\label{code:algosp-delivered} 
	        }
	        \lIf{$v_j = v_0$}{
	            $S \gets \ssuccess$\label{code:algosp-success} 
	        }
	        \If{$B < 0 $}{
	        	\lIf{$flag = \textup{\textbf{true}}$}{
	        		$S \gets \sdelivered$\label{code:algosp-delivery}
	        	} \lElse {
	        		$S \gets \sfail$\label{code:algosp-fail} 
        		}
	        }
        	$t \gets t_{v_j}$\;\label{code:algosp-snapshot-update}
	    }
	} \lElse {
			$S \gets \scanceled$\label{code:algosp-abort}
	}
	
	\caption{\algosp\((G, B, \text{destination } v_c)\)}
	\label{alg:algosp}
\end{algorithm}

Initially, the status $S$ is unset and the delivery $flag$ is set to {false} (Line~\ref{code:algosp-status}).
Given the graph $G_0$, the budget \(B\) and a destination vertex $v_c$, \algosp calculates the shortest cycle $C$ (Line~\ref{code:algosp-shortestcycle}) taking into account only the current wind in $G_0$.
In other words, the drone estimates the route only once at the beginning.
If $d_0(C)\le B$, the drone starts and follows $C$. 
However, during the mission the wind varies as indicated in $\tdg$; so, for each edge the amount of energy actually consumed can be different from the amount accounted at $G_0$.
After traversing each edge, the snapshot is updated as $G_{t_{v_j}}$, where $t_{v_j}$ is the current snapshot when the vertex $v_j$ is reached
(Line~\ref{code:algosp-snapshot-update}).
In this way, the residual energy budget is computed according to the actual edge cost.
In this algorithm, we distinguish the status among \ssuccess, \sfail, and \sdelivered.
\algosp also sets a policy such that if the estimated budget at the beginning is larger than $B$, the mission is canceled with a final status \scanceled (Line~\ref{code:algosp-abort}).

Given a destination, the shortest cycle can be computed (Line~\ref{code:algosp-shortestcycle}) by implementing Dijkstra's algorithm.
Therefore, the \algosp algorithm has a time complexity of \(\mathcal{O}(|E| + |V|\log |V|)\), considering an implementation with an efficient priority queue data-structure, such as Fibonacci Heap.

\paragraph{Example:}
Let us illustrate algorithm \algosp using Figure~\ref{fig:graph-example3}.
It computes the path $(s,a,b,c,d,c,b,a,s)$ at snapshot $G_0$ when the cost is $74$.
However, after performing the cycle $C$, the cost becomes $35$ to reach $d$ and it is $76$ overall to return to the depot.
So, for $B < 74$, the mission returns \scanceled; for $B = 74$, the mission returns \sfail; and for $74 \le B < 76$, the mission returns \sdelivered;
otherwise it returns \ssuccess.
For the example in Figure~\ref{fig:graph-example4}, recalling that the optimal solution has cost $42$ traversing $(s,a,b,d,b,a,s)$, invoking the \algosp algorithm at snapshot $G_0$, we get the shortest-path $(s,a,b,d,b,a,s)$ with cost $18$. 
However, while the drone flies, it spends energy $42$.
So, if the budget is $ B < 18$, the mission returns \scanceled.
If $18 \le B < 42$, the mission returns \sdelivered.
Whereas, if $B \ge 42$, the mission returns \ssuccess.

\subsection{The \algdsp Algorithm}
The \algdsplong (\algdsp) is an online algorithm that dynamically recomputes the shortest path from each intermediate vertex to the destination. 
Here no check is done on the cost of the entire path with respect to the residual budget. 
The recomputed path could be infeasible for the current budget, but the drone moves in any case because it is the path of minimum cost and any other path will cost more than the one just recomputed.

Algorithm \algdsp (reported in Algorithm~\ref{alg:algdsp}) works as follows.
Initially, the current status $S$ is unset, the first snapshot $t=0$ is set, the cycle $C$ does not contain edges, and the current source is $v_i=v_0$ (Line~\ref{code:algdsp-init}).
Then, the algorithm starts to repeat the main loop until the final status is declared (Line~\ref{code:algdsp-stop}).
The first step is to calculate the shortest path $\pi_{ic}$ in \(G_t\) from the current vertex in which the drone resides (i.e., $v_i$) to the current destination, i.e., $v_c$ (Line~\ref{code:algdsp-shortestpath}).
The drone will visit the first edge of such a path (Line~\ref{code:algdsp-firstedge}).
The remaining budget $B$ and current cycle $C$ are updated (Line~\ref{code:algdsp-add}).
Then, traversed the edge, the snapshot is updated (Line~\ref{code:algdsp-snapshot-update}).
The residual graph is also updated removing the vertex $v_i$ and its edges (Line~\ref{code:algdsp-newgraph}), followed by the update on the currently visited new vertex (Line~\ref{code:algdsp-updatecurrent}).
This step is fundamental to avoid the drone to loop endlessly and never reach the destination.
Nevertheless, this step could also lead to failing the mission.
The aforementioned loop is done until either the drone reached the customer or the depot.
However, if for any reason the current budget is exhausted during the mission, a failure is declared (Line~\ref{code:algdsp-fail}).

Similar to \algosp, the time complexity of the \algdsp algorithm is \(\mathcal{O}(|C| \cdot (|E| + |V|\log |V|))\) since for each step of the cycle we have to recompute a new shortest path.

\paragraph{Example:}
Referring to the example in Figure~\ref{fig:graph-example3}, the \algdsp algorithm computes the solution at $G_0$ which is $(s,a,b,c,d,c,b,a,s)$, and proceeds until it reaches vertex $c$ paying a cost of $15$. 
Then the shortest path towards $d$ is $(c,d)$ because the vertices $\{s,a,b\}$ have been removed. 
The destination is reached with an overall cost of $35$. From $d$, the entire graph is reactivated, and the shortest path towards $s$ costs $11$. Hence, the trip computed by \algdsp costs $46$.    
Therefore, for $B < 35$, the mission returns \sfail; for $B=35$, the mission returns \sdelivered; and for $B \ge 46$ \ssuccess is returned.
As regards to the example in Figure~\ref{fig:graph-example4}, in the \algdsp algorithm the drone moves in $a$ since the shortest path at snapshot $G_0$ is $(s,a,b,d)$.
At time $1$, the shortest-path is recomputed, but since the vertex $s$ cannot be selected, the shortest-path is $(a,b,d)$ and so the drone moves to $b$ and then to $d$. 
Then all the vertices and edges of $G$ are reactivated. 
At $G_3$, the shortest path towards $s$ is $(d,f,e,s)$ which remains as the unique path until the end.

\begin{algorithm}[ht]
    $S \gets \emptyset, flag \gets \textbf{false}, C \gets \emptyset, t \gets 0, v_i \gets v_0$\;\label{code:algdsp-init} 
    \While{$S = \emptyset$} {\label{code:algdsp-stop} 
        $\pi_{ic} \gets \textsc{shortestPath}(G_t, v_i, v_c)$\;\label{code:algdsp-shortestpath} 
        $e=(v_i,v_j) \gets \text{first edge of }\pi_{ic}$\;\label{code:algdsp-firstedge} 
        $B \gets B - d_t(e), C \gets C \cup e$\;\label{code:algdsp-add}
        $V \gets V \setminus \{v_i\}, E \gets E \setminus Adj(v_i)$\;\label{code:algdsp-newgraph} 
        $t \gets t_{v_j}$\;\label{code:algdsp-snapshot-update}
        $v_i \gets v_j$\;\label{code:algdsp-updatecurrent}
        \If{$B < 0 $}{
        	\lIf{$flag = \textup{\textbf{true}}$}{
        		$S \gets \sdelivered$\label{code:algdsp-delivery}
        	} \lElse {
        		$S \gets \sfail$\label{code:algdsp-fail} 
        	}
        }
        \If{$v_j = v_c$}{\label{code:algdsp-oneedge} 
            \If{$flag = \textup{\textbf{false}}$}{
                $flag \gets \textbf{true}$\;\label{code:algdsp-delivered} 
        		$(V,E) \gets (G.V, G.E),\; v_i \gets v_c, v_c \gets v_0$\;
            } \lElse{
                $S \gets \ssuccess$\label{code:algdsp-success} 
            }
        }
    }

	\caption{\algdsp\((G,B, \text{destination } v_c)\)}
	\label{alg:algdsp}
\end{algorithm}

\subsection{The \alggsp Algorithm}
The \alggsplong (\alggsp) algorithm,  reported in Algorithm~\ref{alg:alggsp}, is a very short-sighted algorithm that computes the cycle $C$ considering only the most favorable edge at the current time-slot, instead of the entire shortest path from the current vertex to the destination as computed by algorithm~\algdsp.
In particular, at each time-slot \(t\), the strategy is simply to reach the neighbor vertex with the least edge cost considering graph \(G_t\).
There is no guarantee that \alggsp moves in the direction to reach the destination, but it is locally very thrifty. 
Similar to the \algdsp, in this case also we forbid the drone to consider already visited vertices as a next step. 
The mission is declared \sfail if the residual budget is not enough to traverse any outgoing edge, or because the drone cannot continue its path which may occur if the current vertex has no outgoing edges. 

\begin{algorithm}[ht]
    $S \gets \emptyset, flag \gets \textbf{false}, C \gets \emptyset, t \gets 0, v_i \gets v_0$\;\label{code:alggsp-init} 
    \While{$S = \emptyset$} {\label{code:alggsp-stop} 
        $e=(v_i,v_j) \gets \argmin_{k \in Adj(v_i)} d(v_i,v_k)$\;\label{code:alggsp-edge} 
        \lIf{$e = \emptyset$}{
    		$S \gets \sfail$\label{code:alggsp-fail}
    	} \Else {
            $B \gets B - d_t(e), C \gets C \cup e$\;\label{code:alggsp-add}
            $V \gets V \setminus \{v_i\}, E \gets E \setminus Adj(v_i)$\;\label{code:alggsp-newgraph} 
            $t \gets t_{v_j}, v_i \gets v_j$\;\label{code:alggsp-updatecurrent}
            
            \If{$B < 0 $}{\label{code:alggsp-statuses}
            	\lIf{$flag = \textup{\textbf{true}}$}{
            		$S \gets \sdelivered$
            	} \lElse {
            		$S \gets \sfail$
            	}
            }
            \If{$v_j = v_c$}{
            	\If{$flag = \textup{\textbf{false}}$}{
            		$flag \gets \textbf{true}$\;
            		$(V,E) \gets (G.V, G.E),\; v_i \gets v_c, v_c \gets v_0$\;
            	} \lElse{
            		$S \gets \ssuccess$ 
            	}
            }
        }
    }

	\caption{\alggsp\((G,B, \text{destination } v_c)\)}
	\label{alg:alggsp}
\end{algorithm}

We remark that the difference between algorithms \algdsp and \alggsp is that the latter selects only the local edge of minimum cost with respect to the current time-slot (Line~\ref{code:alggsp-edge}), while the former selects the shortest path from the current vertex to the destination.
The former, however, utilize only the first edge of the computed shortest path.
If there is no available edge to visit, the mission is declared \sfail (Line~\ref{code:alggsp-fail}).

Since at each step we have to recompute the minimum outgoing edge from the current vertex, the time complexity of \alggsp is \(\mathcal{O}(\max_{v \in V} |Adj(v)| \cdot |C|)\). 
Without knowing \(\mathcal{G}\), it is not possible to pre-compute this minimum, since the cost of each edge may change as the wind changes. 

\paragraph{Example:}
For the examples in Figure~\ref{fig:graph-example3} and Figure~\ref{fig:graph-example4}, the \alggsp algorithm returns the same solution as \algdsp. 

		
		
		

\section{Synthetic-data Performance Evaluation}\label{sec:simulations}
\revision{In this section, we evaluate the performance of our proposed algorithms on a set of synthetic data.}
We generate at random undirected graphs $G$ with $n=25+1$ vertices in the interval $[0, 2] \unit{km^2}$, modeled according to the \er (ER) model to represent the delivery area and routes. 
ER graphs are characterized by the probability $p(n)$ of the existence of edges. 
\revision{The number of edges $m$ in an ER graph is a random variable with expected value $\binom{n}{2}p(n)$.
Depending on the probability $p(n)$, different layouts can be obtained.
When $p(n)$ is low, many clusters of few vertices interconnected by long-isolated edges are created, while when $p(n)$ is high, a giant cluster is built with many sparse vertices at the border.
In the previous case, the ER graph can model small towns connected by isolated roads, while in the latter case, it can characterize a large city~\cite{erdHos1960evolution}.}
In our evaluation, we adopt $p(n) = c \frac{\log n}{n}$,  for a specific constant $c>0$.
The probability that the created ER graph is connected tends to $0$ if $c < 1$ and to $1$ if $c > 1$~\cite{durrett2007random}.
Moreover, we discard disconnected graphs.
After having generated an edge $e$, we split it into two edges $e$ and $e'$ with opposite directions (but same length), thus making $G$ a directed graph. 
Hence, if a drone can go from $v_i$ to $v_j$, it can also go from $v_j$ to $v_i$. 
Finally, once the global wind is randomly generated, the energy cost is computed as explained in Section~\ref{sec:wind}.

Note that we commensurate the width of the delivery area to the drone battery, as we will point out later looking at the number of \vblack vertices.
Moreover, we have imagined $25$ crossing points, which we believe is reasonable in $4 \unit{km^2}$ of a moderate-dense urban area, like a middle city.
\revision{However, note that our algorithms study point-to-point itineraries, so the network density is not so important, but it was mainly a parameter to generate the ER graphs.}

We consider four different ER layouts (scenarios) by varying the value of \(c= 0.5, 1, 1.5, 2\)  and generating $50$ different random graphs for each $c$.
Notice that for smaller values of $c$, the graphs result in longer graph diameter, and also smaller average vertex-degree.
When $c$ increases, the graphs become denser with a shorter diameter and larger degree.
\revision{For example, when $c=0.5$, the average diameter and degree are $8.5$ and $2.5$, respectively, whereas when $c=2$, the average diameter and degree are $3.5$ and $6.0$, respectively.}
An interesting way to interpret the importance of $c$ in our results is as follows. 
A mission in a delivery network with a small diameter ($c > 1$) lasts for a few snapshots, and it is considered a {\em short} mission. 
A mission with a large diameter ($c <1$) lasts for several snapshots, and it is considered a {\em long} mission.
Obviously, the wind conditions have more chances to change during long missions than during short missions.
So, we expect that \algosp works better when the mission is short than when it is long.
Similarly, \algdsp faces better long missions.

As regards to the drone's parameters, we set a single speed $\vdrone = 20 \unit{m/s}$, and for the deliveries we use a maximum allowable payload of $7 \unit{kg}$ for a decent octocopter~\cite{stolaroff2018energy}.
The drone has also a maximum energy budget of $B = 5000 \unit{kJ}$.
For the wind parameters, we uniformly generate at random the global wind speed $\omega_s$ assuming only four distinct values, i.e., $0 \unit{m/s}$ (calm), $5 \unit{m/s}$ (light), $10 \unit{m/s}$ (moderate), and $15 \unit{m/s}$ (strong) which corresponds to the usual speed winds reported in weather forecast.
Global wind direction $\omega_o$ is generated according to the uniform distribution from $0^{\circ}$ to $359^{\circ}$.
Note that we generate a single global wind every time-slot.
Recall that for the relative wind direction $\omega_o(e)$, we simplified the occurrences in four classes defined by
$\omega_o(e) = 0^{\circ}$ (tail),
$\omega_o(e) = 45^{\circ}$ (semi-tail),
$\omega_o(e) = 135^{\circ}$ (semi-head), and
$\omega_o(e) = 180^{\circ}$ (head).
%



\subsection{Pre-Processing Results}
Figures~\ref{fig:colors_c0.5_n25}--\ref{fig:colors_c2.0_n25} plot the results of our \algpplong (\algpp) algorithm by varying the constant \(c\) and the budget $B$.
The number of \vblack vertices is large only for very small budgets, which means that in general the entire area can be served by the drone under favorable wind conditions.
This means the size of the area is adequate for the size of the battery.
As expected, increasing the budget also increases the number of \vgreen vertices which can be served whatever the wind is and for both long and short missions.
The value of the budget for which the number of \vgreen vertices outdoes the number of \vgray vertices decreases when $c$ increases. 
That is, the short missions require less energy than the long ones.
We explain this with a generalization of the triangular inequality because in $G$ the costs are Euclidean (i.e., the costs are distances scaled all by the same constant). 
Our missions are point-to-point: they connect the depot with the customer. 
It is well known that the shortest way to connect the depot to the customer is to trace a straight line. 
\revision{When instead of going directly we fix an intermediate point (outside the segment), we know that the trip that passes for such an intermediate point is longer than the straight line.}
This reasoning can be inductively repeated increasing the number of intermediate points, and always obtaining paths longer than the previous ones.
When the mission is long, the same two points are possibly connected by a path much longer than that of a short mission.


\subsection{Performance of Proposed Algorithms}
Fixing the value of $c$, we test our algorithms by considering one mission for each \vgray vertex of the $50$ random graphs generated for that $c$. 
For a given algorithm, we vary the budget $B$ (battery $100\%$ fully charged) and plot the percentage of \scanceled, \sfail, \sdelivered, and \ssuccess recorded in all the $50$ graphs generated for any $c$ value.

\begin{figure*}[ht]
	\subfloat[\algpp, $c=0.5$.]{%
		\includegraphics[scale=0.79]{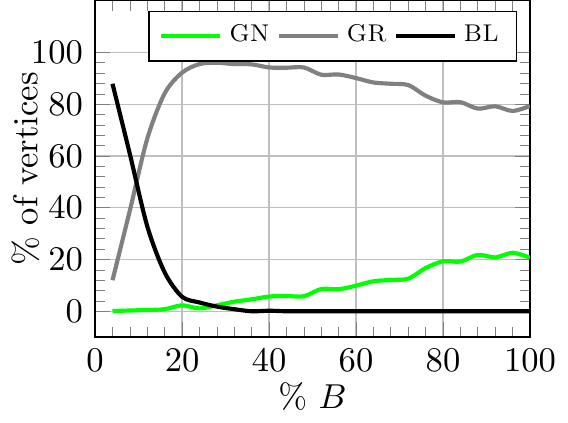}
		\label{fig:colors_c0.5_n25}
	}
	\subfloat[\algpp, $c=1$.]{%
		\includegraphics[scale=0.79]{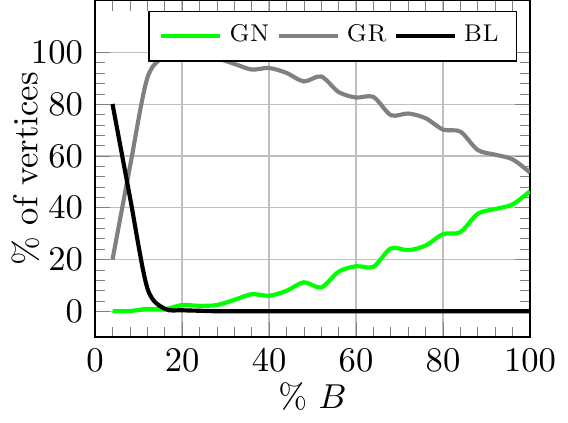}
		\label{fig:colors_c1.0_n25}
	}
	\subfloat[\algpp, $c=1.5$.]{%
		\includegraphics[scale=0.79]{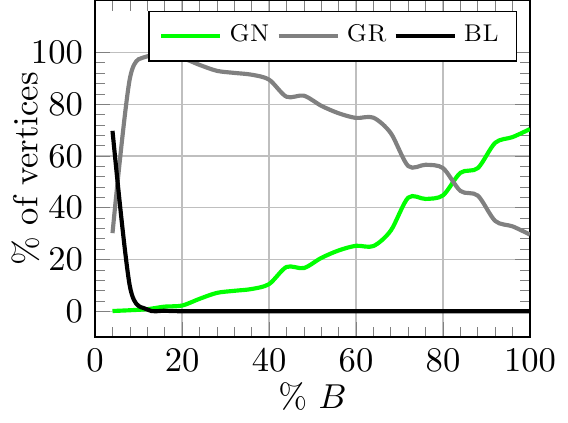}
		\label{fig:colors_c1.5_n25}
	}
	\subfloat[\algpp, $c=2$.]{%
		\includegraphics[scale=0.79]{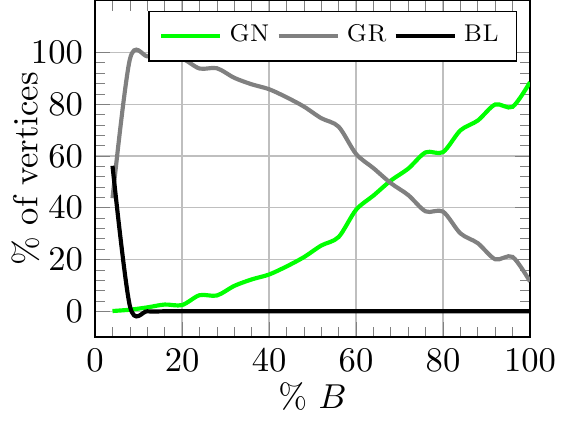}
		\label{fig:colors_c2.0_n25}
	}
    
    \vspace{-0.1in}
    \hfill
    
    \subfloat[\algosp, $c=0.5$.]{%
		\includegraphics[scale=0.79]{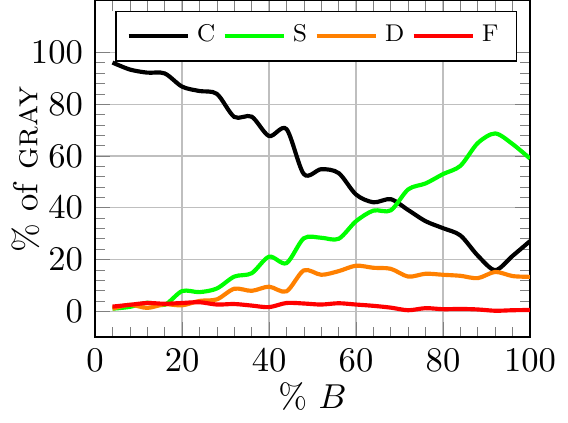}
		\label{fig:alg_off_sp_c0.5_n25}
	}
	\subfloat[\algosp, $c=1$.]{%
		\includegraphics[scale=0.79]{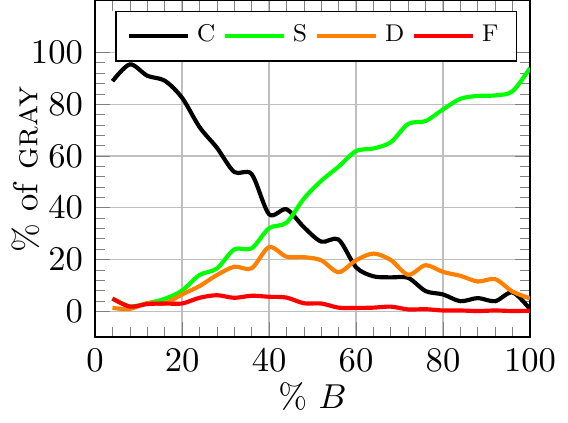}
		\label{fig:alg_off_sp_c1.0_n25}
	}
	\subfloat[\algosp, $c=1.5$.]{%
		\includegraphics[scale=0.79]{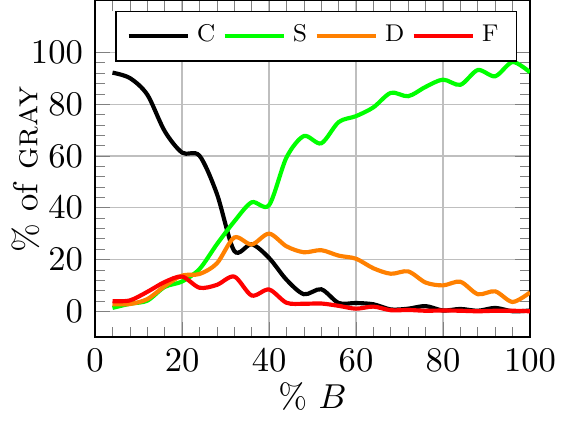}
		\label{fig:alg_off_sp_c1.5_n25}
	}
	\subfloat[\algosp, $c=2$.]{%
		\includegraphics[scale=0.79]{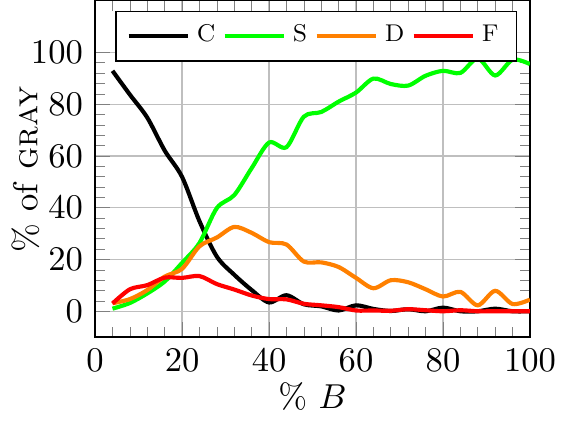}
		\label{fig:alg_off_sp_c2.0_n25}
	}
    
    \vspace{-0.1in}
    \hfill
    
    \subfloat[\algdsp, $c=0.5$.]{%
		\includegraphics[scale=0.79]{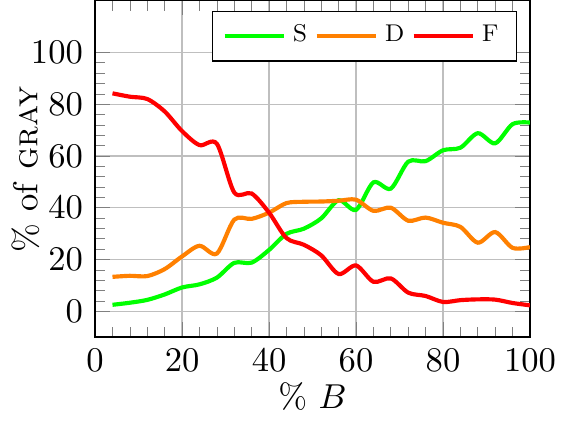}
		\label{fig:alg_on_sp_c0.5_n25}
	}
	\subfloat[\algdsp, $c=1$.]{%
		\includegraphics[scale=0.79]{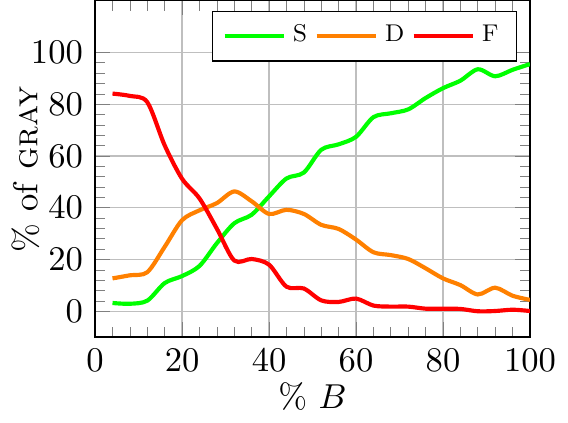}
		\label{fig:alg_on_sp_c1.0_n25}
	}
	\subfloat[\algdsp, $c=1.5$.]{%
		\includegraphics[scale=0.79]{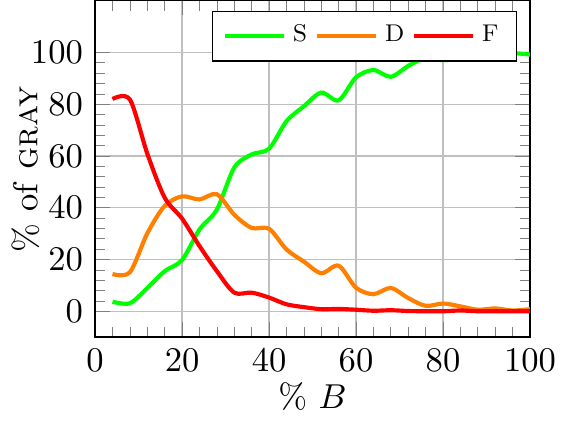}
		\label{fig:alg_on_sp_c1.5_n25}
	}
	\subfloat[\algdsp, $c=2$.]{%
		\includegraphics[scale=0.79]{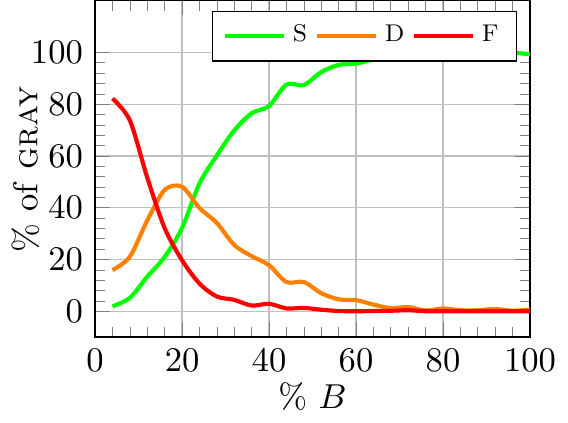}
		\label{fig:alg_on_sp_c2.0_n25}
	}
    
    \vspace{-0.1in}
    \hfill
    
    \subfloat[\alggsp, $c=0.5$.]{%
		\includegraphics[scale=0.79]{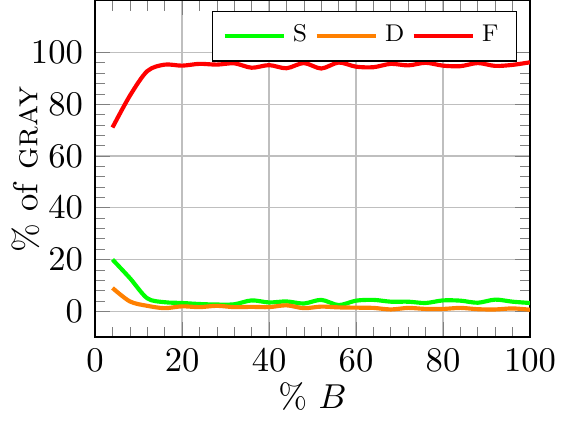}
		\label{fig:alg_on_g_c0.5_n25}
	}
	\subfloat[\alggsp, $c=1$.]{%
		\includegraphics[scale=0.79]{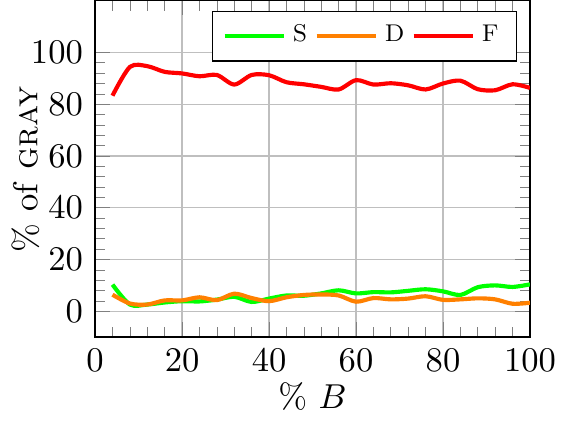}
		\label{fig:alg_on_g_c1.0_n25}
	}
	\subfloat[\alggsp, $c=1.5$.]{%
		\includegraphics[scale=0.79]{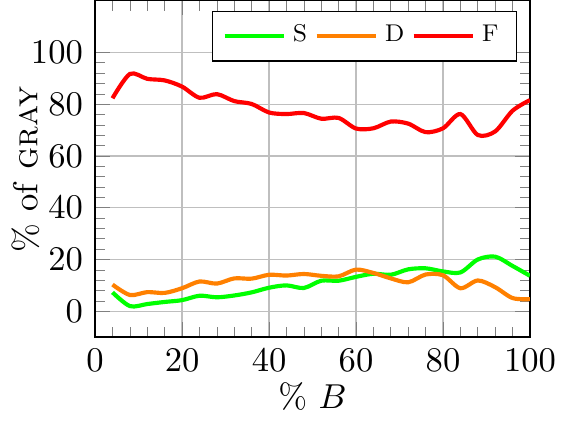}
		\label{fig:alg_on_g_c1.5_n25}
	}
	\subfloat[\alggsp, $c=2$.]{%
		\includegraphics[scale=0.79]{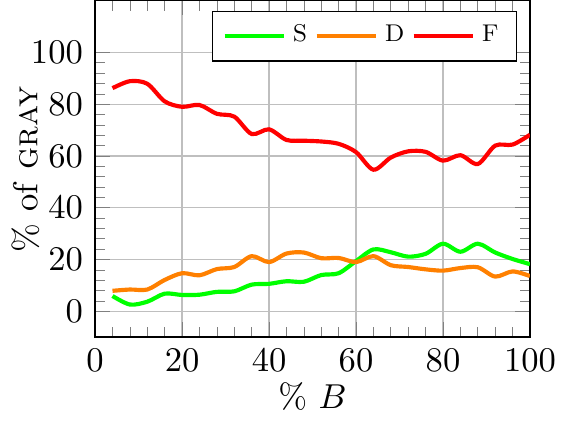}
		\label{fig:alg_on_g_c2.0_n25}
	}
	
	\vspace{-0.1in}
    
	\caption{The colors of vertices for different values of $c$ and budget \(B\) after the \algpplong (\algpp) algorithm (a)--(d), with GN, GR, and BL represent \vgreen, \vgray, and \vblack, respectively.
    Results of 
    \algosplong (\algosp) (e)--(h), 
    \algdsplong (\algdsp) (i)--(l),
    \alggsplong (\alggsp) (m)--(p),
    for different values of \(c\) and budget \(B\),
    with C, S, D, and F represent \scanceled, \ssuccess, \sdelivered, and \sfail, respectively.}
	\label{fig:eeeeeee}
\end{figure*}

Figures~\ref{fig:alg_off_sp_c0.5_n25}--\ref{fig:alg_off_sp_c2.0_n25} show the result of Algorithm~\algosplong (\algosp).
Recall \algosp selects the shortest cycle under the starting wind condition $G_0$, and tries to follow that cycle up to completing the mission.
\algosp is  conservative: if the budget is not sufficient to complete the mission under the condition $G_0$, the mission is \scanceled. 
This approach is reasonable if, based on the weather forecast, $G_0$ is the frequent snapshot in the area. Otherwise, \algosp pays a large number of \scanceled that is not fully justified. This negative effect of \algosp is emphasized when the mission is long (small $c$).
Since the \scanceled curves converge when $c$ is large, we conclude that the conservative approach stops fewer short missions than long ones.
As expected, in \algosp, among the undertaken missions, the percentage of \ssuccess is very high when the mission is short.
The percentage of \sfail increases when the budget is moderate and the mission is short.
\revision{We can explain this by looking at the length of the edges.}
When the missions are short, any point of the delivery map is reached with a few, but long hops.
If the unitary energy efficiency $\mu(e)$ increases, the cost variation is more relevant because $\lambda(e)$ is large. 
Therefore, we suspect that missions that at time $G_0$ were supposed to consume almost all $B$, have more difficulties to tolerate a cost variation on longer edges than shorter ones.


Figures~\ref{fig:alg_on_sp_c0.5_n25}--\ref{fig:alg_on_sp_c2.0_n25} demonstrate the result of Algorithm~\algdsplong (\algdsp).
The \algdsp algorithm, which is computationally the most expensive, yields a larger percentage of \ssuccess than \algosp.
In \algdsp, when the drone is charged at about $30\%$ and the missions are short, the percentage of \ssuccess is higher than $70\%$; whereas, with the same amount of energy, \algosp is able to accomplish only $40\%$ of the missions.
So, \algdsp requires less energy to accomplish a mission.
Nonetheless, the percentage of \sfail is also higher in \algdsp than \algosp.
With a small budget, most of the missions either return \sfail or \sdelivered.
It seems that those missions, \scanceled in \algosp, cannot be completed in \algdsp either. 
Summing the \sdelivered and \ssuccess, even with a minimal budget between $20\%$ and $30\%$, \algdsp serves about $40\%$ of the customers.
In conclusion, the performance of \algdsp is quite good, which will be
even clearer when compared with that of \alggsp. 


Figures~\ref{fig:alg_on_g_c0.5_n25}--\ref{fig:alg_on_g_c2.0_n25} depict the result of Algorithm~\alggsplong (\alggsp), which requires the minimum information about the energy cost of the edges.
Its performance does not depend on the budget because most often the drone does not continue with the online strategy since it reaches a dead end. 
This is probably because the local choice creates holes in the graph and disconnects it.
\alggsp vaguely moves the drone and drains its energy.
When the mission is short, the percentage of \sfail decreases and the local saving due to the short-sighted algorithm alleviates the poor performance. 
In this case, about $40\%$ of the missions are declared with \ssuccess.

In summary, both \algosp and \algdsp algorithms perform better on short missions than longer ones, but \algdsp achieves a very good percentage of \ssuccess even for long missions when the budget is above $50\%$. 
However, \algdsp copes with wind changes, and \algosp does not.
Finally, \alggsp is the fastest, but its performance is poor.
It could be improved by searching in a sightly larger neighborhood to avoid the path is trapped at dead ends.

\section{\revision{real data Performance Evaluation}}\label{sec:simulations2}
In this section, we evaluate the performance of our proposed algorithms on a set of real data.
We obtained a full data-set of winds registered from $12$ \wcus located in Corsica, France.
This comprehensive one-month long data-set is so-structured: date, time, global wind speed, global wind direction, \wcu location.
Data are updated every $60 \unit{mins}$.

Although this data-set is quite interesting, it is not directly suitable for our context.
The first concern is that the distances between any pair of stations are, on average, too long for the considered octocopter.
The same can be said for the wind data updates frequency, which is too large in terms of time with respect to the octocopter's maximum flight time.
To adjust the real data to our scenario, we have scaled the distances by a factor of $10$, and scaled the time-slots by a factor of $4$, assuming that winds are updated every $15 \unit{mins}$. 
So, from now on, we refer to our data set as \emph{quasi-real data}.
The second concern is how to map any point of the delivery area with a particular \wcu.
For this scope, we tessellate the delivery area according to the \vor tessellation~\cite{van2000computational} built with the \wcus.
Each point of the delivery area is then mapped to the closest (Euclidean distance) \wcu forming \vor-cells, each centered at a different \wcu.
The third concern is about the routes that the drone can follow among the \wcus.
For this, we have built three types of graphs: \vor-graph (VG), \del-graph (DG), and \hyb-graph (HG).
All these graphs include the $12$ \wcus among their vertices.

\subsection{Graphs Construction}
The vertices $V_V$ of the VG are the \wcus and the vertices of the \vor-cells.
The set of edges $E_V$ will be composed of the set of undirected edges that join the vertices of each \vor-cell to the \wcu at its center and all the sides that delimit the \vor-cells.
Each edge is then split into $2$ edges with opposite directions to make the VG directed.
To determine the unitary energy cost function in Eq.~\eqref{eq:unitary-energy}, each edge $e$ is associated with the global wind observed at time $t$ in the \vor-cell where it resides. 
The $2$ directed edges obtained by splitting the side of the \vor-cell are associated with the $2$ global winds of the $2$ adjacent \vor-cells that they delimit. 
Precisely, each directed edge is associated with the wind of the \vor-cell on its right. 
%

The vertices $V_D$ of the DG are the \wcus and some of the vertices of the \vor tessellation that are introduced by building the edges.
The edges $E_D$ are built using the \del triangulation, which represents the dual graph of the \vor tessellation. 
Each edge of the \del triangulation connects $2$ centers of $2$ \vor cells, i.e., $2$ \wcus.
$E_D$ will contain all the edges belonging to the \del triangulation that cross exactly $2$ adjacent \vor cells. 
The \del edges that do not satisfy this property (i.e., they cross multiple \vor cells) are eliminated and replaced by $2$ edges as follows.
Precisely, the eliminated edge that connects the \wcus $v_i$ and $v_j$ is replaced by $2$ edges $(v_i,v_k)$ and $(v_k,v_j)$ that pass through the $v_k$ vertex of the \vor tessellation.
Such a \vor vertex $v_k$ is the vertex that minimizes the Euclidean distances between itself and $v_i$ and $v_j$. 
Clearly,  $v_k$ is inserted in $V_D$.  
To determine $\mu_{\omega}(e)$ in DG, we observe that the edges that were not originally in the \del triangulation belong entirely to single \vor-cell. The other edges, instead, reside for half in the cell of one of their extremes, and the other half in the cell of the other extreme. 
Formally, for $e=(v_i,v_j)$, the Eq.~\eqref{eq:energy} can be rewritten as: $d_\omega(e)=\mu_{\omega(v_i)}\frac{\lambda(e)}{2} + \mu_{\omega(v_j)}\frac{\lambda(e)}{2}=\frac{\mu_{\omega(v_i)}+\mu_{\omega(v_j)}}{2}\lambda(e)$.

The HG is finally constructed from the union of the $2$ previous graphs: $V_H=V_V \cup V_D=V_V$ and $E_H=E_V \cup E_D$.
Each edge inherits the energy cost of the original graph.

In conclusion, all the edges are subjected to at most $2$ winds, i.e., those circumscribed by the regions of \vor that enclose the $2$ centers.
We did not consider the complete Euclidean graph whose edges connect each \wcu to each other since it would have been difficult to determine the energy cost associated with each edge. Namely, each Euclidean edge traverses several \vor-cells for different lengths.
Moreover, the diameter of VGs, DGs, and HGs are $6$, $6$, and $7$, respectively. The DG has a diameter larger than VG and HG observed that DG has $19$ vertices, while the others have $34$ vertices. 

\begin{table}[t]
	\renewcommand{\arraystretch}{1.25}
	\caption{Comparison between \algosp, \algdsp, and \alggsp on quasi-real data. C, S, D, and F represent \scanceled, \ssuccess, \sdelivered, and \sfail, respectively. Data is in percentage.}
	\label{tab:real_data}
	\centering
	\begin{tabular}{p{0.05cm}c|c|cccc|ccc|ccc}

		& & & \multicolumn{4}{|c}{\algosp} & \multicolumn{3}{|c}{\algdsp} & \multicolumn{3}{|c}{\alggsp} \\
        \hline
		
		& $B$ & gray & C & S & D & F & S & D & F & S & D & F \\
		
		\hline
		\multicolumn{13}{c}{4 wind classes} \\
		
		\hline

		\multirow{2}{*}{\rotatebox[origin=c]{90}{VG}} & 100&88	&9	&90	&1	&0	&90	&10	&0	&7	&25	&68\\
		& 50&89	&44	&56	&0	&0	&56	&39	&5	&3	&21	&76\\
		\hline
		
		\multirow{2}{*}{\rotatebox[origin=c]{90}{DG}} & 100&80	&7	&92	&1	&0	&93	&7	&0	&13	&36	&51\\
		& 50&85	&42	&57	&1	&0	&58	&38	&4	&6	&29	&64\\
		\hline
		
		\multirow{2}{*}{\rotatebox[origin=c]{90}{HG}} & 100&80	&3	&97	&0	&0	&97	&3	&0	&13	&31	&56\\
		& 50&86	&34	&66	&0	&0	&66	&33	&1	&6	&27	&67\\
		\hline

		
		\multicolumn{13}{c}{8 wind classes} \\
		
		\hline
		
        \multirow{2}{*}{\rotatebox[origin=c]{90}{VG}} & 100 &88 & 8	 &91 &1	&0	& 92 & 8	& 0	& 7	 &23	 & 71 \\
        & 50 &89 & 42	 &58 &0	&0	& 58 & 40	& 2	& 3	 &21	 & 76 \\
        \hline
        
        \multirow{2}{*}{\rotatebox[origin=c]{90}{DG}} & 100 &80 & 3	 &96 &1	&0	& 97 & 3	& 0	& 19 &38	 & 44 \\
        & 50 &85 & 35	 &65 &1	&0	& 65 & 34	& 1	& 8	 &32	 & 58 \\
        \hline
        
        \multirow{2}{*}{\rotatebox[origin=c]{90}{HG}} & 100 &80 & 2	 &98 &0	&0	& 98 & 2	& 0	& 18 &30	 & 52 \\
        & 50 &86 & 25	 &75 &0	&0	& 75 & 25	& 0	& 8	 &26	 & 66 \\
		\hline
	\end{tabular}
\end{table}

\subsection{Performance of Proposed Algorithms on real data}
In this section, we present the results for the quasi-real data.
In the evaluation, we assume to use the same octocopter as we did in Section~\ref{sec:simulations}. 
However, for the quasi-real tests, we assume the same battery budget ($B=5000 \unit{kJ}$) and report the results with the $100\%$ and the $50\%$ of $B$. 
This time, the drone flies at a slower speed ($\vdrone = 10 \unit{m/s}$) and carries less payload ($2 \unit{kg}$). 
Eq.~\eqref{eq:unitary-energy} is computed assuming $4$ or $8$ wind classes, i.e., the representative winds are $i \cdot 45^{\circ}$, with $i=0, \ldots, 3$ and  $i \cdot 22.5^{\circ}$, with $i=0, \ldots, 7$, respectively.

The results relative to the mission statuses are reported in Table~\ref{tab:real_data}. 
We first note that the \alggsp algorithm performs very poorly under all the conditions: it has a large percentage of \sfail and a very small percentage of \ssuccess. Moderate is the percentage of \sdelivered. 
\algosp and \algdsp algorithms have a similar percentage of \ssuccess status, despite the higher computation complexity and the higher information required at each new reached vertex.
However, it is worth noticing that most of the missions that \algosp consider as \scanceled are declared as \sdelivered by the \algdsp algorithm. 
The passage from $4$ to $8$ wind classes does not bring particular benefits. 
Only \algdsp improves the success rate for DGs showing that in DGs the angle directions $\psi(e)$ are more varied. This is interesting, but not surprising because the \del triangulation aims to maximize the width of the minimum angle in each built triangle.
As expected, the HG is the graph that yields a higher percentage of \ssuccess because it has more routes than the other graphs.
The VG is the scenario in which the algorithms perform slightly worse, probably because the \vor edges are longer than \del edges, which are at most $3$ times the Euclidean distance between their endpoints~\cite{keil1989delaunay}.
Moreover, we can notice that, when halving the battery, even if the percentage of \vgray nodes does not change much, it is the number of \vgreen nodes that decreases. 
Moreover, the number of \ssuccess missions is almost halved in all the tests.
Considered that the VG and the other $2$ graphs can be compared with a synthetic graph with $c=0.5$ and $c=1.5$ respectively, the results on the quasi-real data are high likely those obtained on the synthetic data in Section~\ref{sec:simulations}.

In conclusion, our tests support the commercial benefit of wind awareness when planning a mission. We show that a very conservative approach that only starts a mission if the initial wind conditions allow us to complete it, like \algosp does, cancels up to $9\%$ of the deliveries if the routes are long. While a more risky approach, such as \algdsp, requires more control information but completes all the deliveries and successfully returns the drone to the depot in more than $90\%$ of the missions. On the other hand, a short-sight strategy that considers only the local wind conditions, like \alggsp, can fail more than $50\%$ of the deliveries abandoning the drone somewhere.

\section{\revision{Conclusion}}\label{sec:conclusions}
In this paper, we introduced the \prob which studies the feasibility of sending a drone to deliver goods from the depot to a given customer with a limited energy battery budget.
To properly model the \prob, we proposed a framework based on time-dependent graphs.
We proposed three algorithms, \algosp, \algdsp, and \alggsp, and evaluate their performance in terms of accomplished missions.
Results on synthetic and quasi-real data show that \algosp is the most conservative approach, \algdsp completes more deliveries at a higher risk of failure, and \alggsp is the least performing.
%
As future work, it is reasonable to think about probabilistic approaches based on wind forecasts, to guess the most likely wind conditions during the mission.
Moreover, in our proposed model some  improvements have still to be addressed, for example, consider going back at the depot once the drone has an unexpected battery consumption, or stop at specific positions waiting for more favorable wind conditions.

\paragraph{Acknowledgments}
The authors are grateful to the editor and reviewers for valuable comments that helped us improve the manuscript quality.
The authors are grateful to Patric Botey, director of  R\&D Projets Europeens Civil Protection at  SDIS, and to Florence Vayess, Référente Nationale of Meteo-France for providing real data.
Finally, the authors thank Lorenzo Palazzetti for developing the tests with the real data.
This work was supported by Intelligent Systems Center (ISC) at Missouri S\&T,
and by NSF grants CNS-1545050, CNS-1725755, SCC-1952045, and Fondo Ricerca di Base, 2019, UNIPG.

\clearpage

\bibliographystyle{IEEEtran}
\bibliography{IEEEabrv,main}

\vspace{-0.4in}
\begin{IEEEbiography}
[{\includegraphics[width=1in,height=1.25in,clip,keepaspectratio]{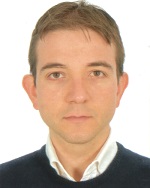}}] {Francesco Betti Sorbelli}
received the Bachelor and Master degrees {\em cum laude} in Computer Science from the University of Perugia, Italy, in 2007 and 2010, respectively, 
and his Ph.D. in Computer Science from the University of Florence, Italy, in 2018.
After his Ph.D., he was a Postdoc Researcher at University of Perugia.
Currently, he is a Postdoc at the
Missouri University of Science and Technology University, USA.
His research interests include wireless sensor networks,
algorithms on drones and robots.
\end{IEEEbiography}

\vspace{-0.3in}
\begin{IEEEbiography} [{\includegraphics[width=1in,height=1.25in,clip,keepaspectratio]{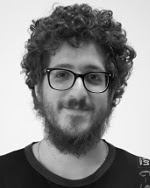}}] {Federico Corò}
received the Bachelor and Master degrees {\em cum laude} in Computer Science from the University of Perugia, Italy, in 2014 and 2016, respectively and his Ph.D. in Computer Science in 2019 at Gran Sasso Science Institute (GSSI), L’Aquila, Italy.
Currently, he is a post-doctoral researcher in the Department of Computer Science at La Sapienza in Rome, Italy.
His research interests include several aspects of theoretical computer science, including combinatorial optimization, network analysis, and the design and efficient implementation of algorithms.
\end{IEEEbiography}

\vspace{-0.2in}
\begin{IEEEbiography} [{\includegraphics[width=1in,height=1.25in,clip,keepaspectratio]{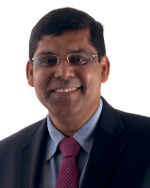}}] {Sajal K. Das} is a professor of computer science and Daniel St. Clair Endowed Chair at Missouri University of Science and Technology. 
His research interests include wireless sensor networks, mobile and pervasive computing, cyber-physical systems and IoT, smart environments, cloud computing, cyber security, and social networks. He serves as the founding Editor-in-Chief of Elsevier's Pervasive and Mobile Computing journal, and as Associate Editor of several journals including the IEEE Transactions of Mobile Computing, IEEE Transactions on Dependable and Secure Computing, and ACM Transactions on Sensor Networks. He is an IEEE Fellow.
\end{IEEEbiography}

\vspace{-0.3in}
\begin{IEEEbiography} [{\includegraphics[width=1in,height=1.25in,clip,keepaspectratio]{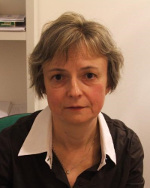}}] {Cristina M. Pinotti}
received the Master degree {\em cum laude} in Computer Science from the University of Pisa, Italy, in 1986.
In 1987-1999, she was Researcher with the National Council of Research in Pisa.
In 2000-2003, she was Associate Professor at the University  of Trento.
Since 2004, she is a Full Professor at the University of Perugia.
Her current research interests include the design and analysis of algorithms for wireless sensor networks and communication networks.
\end{IEEEbiography}

\end{document}